\documentclass[journal]{IEEEtran}

\ifCLASSINFOpdf
\else
   \usepackage[dvips]{graphicx}
\fi
\usepackage{url}

\usepackage{graphicx}
\usepackage{amsmath}
\usepackage{multirow}
\usepackage{dblfloatfix}
\usepackage{float}
\usepackage{algorithm}
\usepackage{subcaption}
\usepackage{algpseudocode}
\usepackage{float}
\usepackage{titlesec}

\usepackage{hyperref}
\usepackage{caption}

\begin{document}

\title{
    {\small \textbf{This work has been submitted to the IEEE for possible publication. Copyright may be transferred without notice, after which this version may no longer be accessible.}}\\
    $S^2IL$: Structurally Stable Incremental Learning
}

\author{S Balasubramanian, Yedu Krishna P, Talasu Sai Sriram, M Sai Subramaniam, \\Manepalli Pranav Phanindra Sai, Darshan Gera

\centering{Sri Sathya Sai Institute of Higher Learning}
% \IEEEpubid {This work has been submitted to the IEEE for possible publication. Copyright may be transferred without notice, after which this version may no longer be accessible.
% }
\thanks{Date of submission: 20th March, 2025}

\thanks{All the authors are affiliated to the Department of Mathematics and Computer Science (DMACS), Sri Sathya Sai Institute of Higher Learning, Prasanthi Nilayam, India.}
}

% \markboth{IEEE Signal Processing Letters, March 2025}
% {Shell \MakeLowercase{\textit{et al.}}: }
% \twocolumn[{
%     \centering
%     {\small This work has been submitted to the IEEE for possible publication. Copyright may be transferred without notice, after which this version may no longer be accessible.\par}
%     \vspace{1em} % Add some vertical space before the title
%     % \maketitle
% }]
\maketitle

\begin{abstract}
Feature Distillation (FD) strategies are proven to be effective in mitigating Catastrophic Forgetting (CF) seen in Class Incremental Learning (CIL). However, current FD approaches enforce strict alignment of feature magnitudes and directions across incremental steps, limiting the model's ability to adapt to new knowledge. In this paper we propose Structurally Stable Incremental Learning($S^2IL$), a FD method for CIL that mitigates CF by focusing on preserving the overall spatial patterns of features which promote flexible (plasticity) yet stable representations that preserve old knowledge (stability). We also demonstrate that our proposed method $S^2IL$ achieves strong incremental accuracy and outperforms other FD methods on SOTA benchmark datasets CIFAR-100, ImageNet-100 and ImageNet-1K. Notably, $S^2IL$ outperforms other methods by a significant margin in scenarios that have a large number of incremental tasks. \footnote{Code will be released on acceptance of the paper.}
\end{abstract}


\IEEEpeerreviewmaketitle

\section{Introduction}

\IEEEPARstart{H}{umans} continuously learn, retaining knowledge from past experiences. Efforts aim to enable machines to mimic this for solving real-world problems \cite{Gupta_2023_CVPR, Park_2021_ICCV}. Class Incremental Learning (CIL) \cite{Kuzborskij_2013_CVPR, Li_2016_ECCV, Rebuffi_2017_CVPR, Hou_2019_CVPR, Liu_2021_CVPR, Kim_2023_CVPR, He_2024_CVPR}, where models adapt to new classes while retaining knowledge of old ones, has gained attention. The success of deep neural networks (DNNs) has led to various approaches to CIL \cite{Yan_2021_CVPR, Wen_2024_CVPR, Zenke_2017_ICML, Kang_2022_CVPR}, mainly focusing on mitigating catastrophic forgetting (CF), where adapting to new classes causes loss of previous knowledge. A key approach is feature distillation (FD), which aligns features across tasks to preserve representations of old classes while training on new ones. SOTA FD methods typically minimize the squared-norm \cite{Hou_2019_CVPR, Dou_2020_ECCV}, or weighted squared-norm \cite{Bala_2024_arXiv, Kang_2022_CVPR}, of the difference between feature maps from current and past tasks, with weights reflecting feature importance. However, as shown in Figure 1a and Section III-A, importance values often flatten at 1, reducing these methods \cite{Bala_2024_arXiv, Kang_2022_CVPR} to simple squared-norm minimization, enforcing alignment in both magnitude and direction thereby promoting stability but restricting adaptability to new classes (plasticity).

\begin{figure}[H]
    \centering

    % First subfigure
    \begin{subfigure}[b]{0.52\textwidth}  % Adjust width to fit two figures side by side
        \centering
        \includegraphics[width=\textwidth]{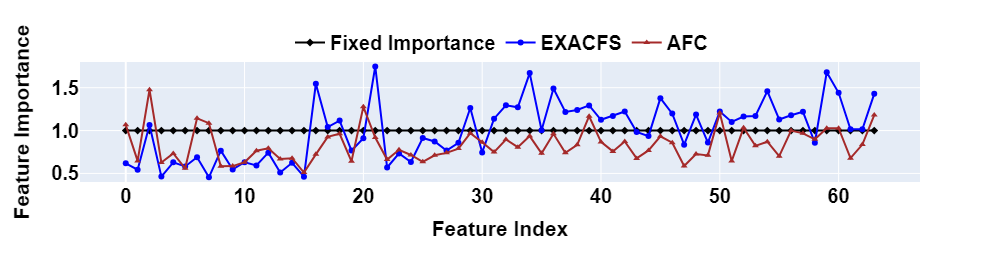}  % Replace with your first image
        \caption{}
        \label{fig:importance_plot}
    \end{subfigure}
    \hfill  % Adds horizontal space between the subfigures

    % Second subfigure
    \begin{subfigure}[b]{0.45\textwidth}  % Adjust width to fit two figures side by side
        \centering
        \includegraphics[width=0.8\textwidth]{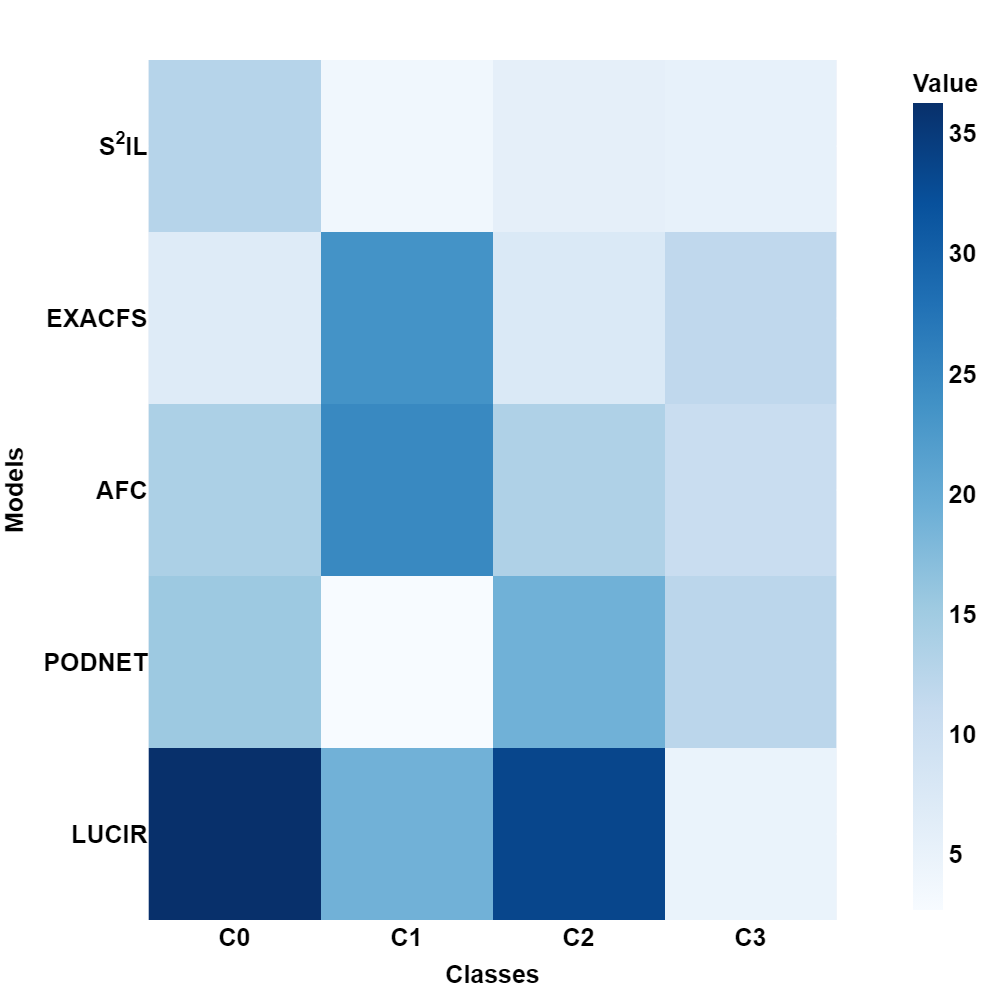}  % Replace with your second image
        \caption{}
        \label{fig:deviation_heatmap}
    \end{subfigure}
    \caption{A motivation for exploring structure based feature distillation (FD): (a) Average feature importance $\rho$ across increments from the last convolutional layer of two SOTA FD models, EXACFS and AFC, evaluated on CIFAR100 with a \textbf{Inc 10} setting. Both enforce feature similarity (magnitude and direction) between increments. A feature is deemed important if it significantly influences the loss. Surprisingly, $\rho$ remains nearly constant at $1$ for all the features, for both models, limiting plasticity by enforcing feature similarity of all corresponding features equally. This highlights the need for a FD idea that balances stability with plasticity. (b) Heatmap comparing class-wise deviations in Grad-CAM feature importances of various models from those of the Oracle model $O$. $O$ is trained like any other CIL model except that it has access to all past and current train data. $S^2IL$ shows significantly lower deviation from $O$, suggesting that accounting for feature structure results in better generalization.}
    \label{fig:mainfig}
\end{figure}

To overcome these limitations, we propose the following contributions:
\begin{itemize}
    \item We interpret the \textbf{Structural Similarity Index Measure (SSIM)} \cite{Wang_2004_TIP} in the CIL setting.
    \item A FD approach called Structurally Stable Incremental Learning ($S^2IL$) that preserves \textbf{structural similarity} of features across incremental steps instead of enforcing exact feature alignment, which encourages the model to retain the overall spatial patterns within features to promote flexible yet stable representations. 
    \item The efficacy of $S^2IL$ is demonstrated on benchmark datasets.
\end{itemize}

\vspace{-2mm}
\section{The CIL Formulation}
\label{sec:formulation}
Assume $T + 1$ tasks with task $0$ being the base task and the remaining tasks $1$ through $T$ arriving incrementally. The dataset for task $t$, $t \in \{0, 1, ..., T\}$, is $D^t = \{(x^t_i, y^t_i)\}_{i=1}^{n^t}$ where $x^t_i$ is the data, $y^t_i$ is the class label and $n^t$ is the number of samples. Each class label $y^t_i$ belongs to $C^t$ where $C^t = \{c^t_1, c^t_2, ...,, c^t_{m^t}\}$ represents the set of $m^t$ classes associated with task $t$. For $i \neq j$, we assume $C^i \cap C^j = \emptyset$. Let $D^{0\sim t} = \cup_{i=0}^{t}D^i$ and $C^{0 \sim t} = \cup_{i=0}^{t}C^i$.  
 Each task $t$ also receives a small set of exemplars $E^t_i$ from each previously seen class $c^t_i$. 
 
Depending on the context, $D^t$ ($D^{0\sim t}$) may refer to either the training or testing data for task $t$ (for tasks $0$ to $t$). 

We assume a fixed convolutional network $F$ appended with a global average pooling layer $G$ and a growing classifier $H$ across tasks. During training on task $t$, $H$ has $m^t$ nodes added to its existing nodes. The network $F$ consists of $L$ layers, with $f_{ij}$ denoting the $j^{th}$ feature map of layer $i$. Let $i_{fm}$ represent the number of feature maps in convolutional layer $i$. Let $f_{(L+1)}$ denote the output of global average pooling layer $G$. 
We denote the overall model by $M = H \circ G \circ F$, where `$\circ$' is the composition operation. At task $t$, we denote $M$, $F$, $G$ and $H$ by $M^t$, $F^t$, $G^t$ and $H^t$, respectively. Similarly, the feature map $f_{ij}$ at task $t$ is denoted by $f_{ij}^t$. During training of $M^t$, the parameters of $F^t$ and the parameters of $H^t$ associated with the previously seen classes are initialized with parameters from the corresponding components of $M^{t-1}$. The new nodes of $H^t$ are initialized using Imprinted Weights \cite{Qi_2018_CVPR} as in \cite{Dou_2020_ECCV}. For $t=0$, the entire $M^0$ is initialized randomly.

\section{The Proposed $S^2IL$} 
\label{sec:s2il}
\subsection{Motivation}
\label{sec:motivation_s2il}
Current SOTA FD approaches \cite{Hou_2019_CVPR, Dou_2020_ECCV, Kang_2022_CVPR, Bala_2024_arXiv}, in a generic sense, minimize one of the following objectives specific to feature preservation during the training of task $t$.
\begin{equation}
    \sum_{i=1}^{L}\sum_{j=1}^{i_{fm}}||f^t_{ij}(x) - f^{t-1}_{ij}(x)||^2 + ||f^t_{L+1}(x) - f^{t-1}_{L+1}(x)||^2
    \label{eq:std_fd}
\end{equation}
\begin{equation}
    \sum_{i=1}^{L}\sum_{j=1}^{i_{fm}}\rho^t_{ij}||f^t_{ij}(x) - f^{t-1}_{ij}(x)||^2 + \rho^t_{L+1}||f^t_{L+1}(x) - f^{t-1}_{L+1}(x)||^2
    \label{eq:imp_fd}
\end{equation}
where $x$ is the input sample, $||.||$ is Frobenius norm and $\rho^t_{ij}$ is the importance of feature map $j$ in layer $i$ of $F^t$ during training of task $t$, which helps preserve the learned model from task $t-1$. $\rho^t_{L+1}$ is similarly defined for features from $G^t$. \cite{Hou_2019_CVPR} considers only the second term in Eqn. \eqref{eq:std_fd} with normalized $f^t_{L+1}$, which simplifies the minimization to the maximization of $<f^t_{L+1}, f^{t-1}_{L+1}>$, where $<. , .>$ denotes the standard inner product in real space. \cite{Dou_2020_ECCV} uses the term corresponding to layer $L+1$ as is, but for the other layers, it considers the squared norm of the difference between the corresponding spatially aggregated features (aggregated separately along width and height) instead of direct comparison. \cite{Kang_2022_CVPR, Bala_2024_arXiv} follow the formulation in Eqn. \eqref{eq:imp_fd}, with \cite{Bala_2024_arXiv} estimating the importance for each layer, feature, and class, while \cite{Kang_2022_CVPR} estimating the importance only for each layer and feature.

Surprisingly, the average importance factor $\rho$ almost flattens to $1$ in both \cite{Bala_2024_arXiv} and \cite{Kang_2022_CVPR} across the layers. This is illustrated in Figure \ref{fig:importance_plot} for the last convolutional layer, which depicts the average $\rho$ value for each feature map, averaged across all incremental tasks on the CIFAR-100 dataset where the base task consists of half the total number of classes, with 10 new classes added per incremental task. The flattening to $1$ was observed in other incremental settings also, as detailed in the supplementary material. In our experiments, the difference in average incremental accuracy (AIA) between setting $\rho$ to $1$ and explicitly estimating $\rho$ is as small as $0.4\%$ in both \cite{Bala_2024_arXiv} and \cite{Kang_2022_CVPR}. In other words, both \cite{Bala_2024_arXiv} and \cite{Kang_2022_CVPR} effectively minimize the squared-norm of the difference between corresponding feature maps, as in Eqn. \eqref{eq:std_fd}, with respect to feature alignment.

A key issue with minimizing terms like $||u-v||^2$ in Eqn. \eqref{eq:std_fd} is that it either aligns the direction of $u$ and $v$ when their sizes are constrained (e.g., normalized) or aligns both their magnitude and direction, as $||u-v||^2 = ||u||^2 + ||v||^2 - 2<u, v>$. Such alignment aims to make the feature representations for a given input nearly identical across tasks, preserving both the direction and magnitude of features in the feature space. However, this strict alignment can hinder the model's plasticity by forcing it to retain exact feature representations, which reduces its capacity to adapt to new classes and patterns. While we present the reasoning for the limitation here, \cite{Kim_2023_CVPR} experimentally validates the lack of plasticity in the approaches of \cite{Hou_2019_CVPR}, \cite{Dou_2020_ECCV}, and \cite{Kang_2022_CVPR}.

The aforementioned limitation motivated us to frame FD as a means of preserving structural similarity between features across tasks. By preserving structure, we aim to maintain the spatial arrangement of features.

This allows for more flexibility in terms of feature magnitudes and directions, providing the model room to adjust these aspects to new data, while still maintaining an overall structural similarity. This is akin to preserving the `shape' rather than exact `positions' in the feature space.  
To enforce structural similarity, we leverage SSIM \cite{Wang_2004_TIP}, and refer to our proposed method as Structurally Stable Incremental Learning ($S^2IL$). 
\vspace{-3mm}
\subsection{The $S^2IL$ Loss}
\label{sec:s2il_loss}
First, we adopt SSIM \cite{Wang_2004_TIP} for a pair of corresponding feature maps $u$ and $v$ between previous and current tasks respectively, and present our interpretations. SSIM between $u$ and $v$ is defined as:
\begin{equation}
\text{SSIM}(u, v) = l(u, v)^p \cdot c(u, v)^q \cdot s(u, v)^r
\end{equation}

where \[
l(u, v) = \frac{2 \mu_u \mu_v + C_1}{\mu_u^2 + \mu_v^2 + C_1}
\quad
c(u, v) = \frac{2 \sigma_u \sigma_v + C_2}{\sigma_u^2 + \sigma_v^2 + C_2}
\]

\[
s(u, v) = \frac{\sigma_{uv} + C_3}{\sigma_u \sigma_v + C_3}
\]

and $p$, $q$, $r$, $C_1$, $C_2$, $C_3$ are positive reals.

$\mu_u$ ($\mu_v$) is the mean activation level of feature map $u$ ($v$) across spatial dimensions.
$\sigma_u^2$ ($\sigma_v^2$) quantifies the variance of feature activations around $\mu_u$ ($\mu_v$). %spatially.
$\sigma_{uv}$ quantifies the linear relationship between the activations of $u$ and $v$, essentially capturing how well the structural patterns of $u$ align with those of $v$.
$l(u, v)$, the luminance component in SSIM, measures the similarity in the mean activations between the two feature maps across tasks. It reflects how well the average activations align spatially, representing their global similarity in terms of the feature values. 
$c(u, v)$, the contrast component, measures the similarity in the activation distributions between the two feature maps across the tasks.
$s(u, v)$, the structure component, measures the similarity in the spatial relationships between the two feature maps across tasks. This captures how well the spatial patterns in $u$ and $v$ align, independent of their magnitudes.

$l$ and $c$ help maintain the global features and the distribution of features across tasks. This means that the model keeps a consistent representation of the features it has learned, even as it adapts to new tasks, without overfitting to old tasks. $s$ ensures that as the model learns new tasks, the spatial structure of the features it learned from previous tasks is retained. This allows the model to reorganize and adjust its internal representations to incorporate new classes or tasks, without completely distorting the learned patterns from old tasks.

Given this adoption of SSIM to CIL setting, we define the FD loss during training of task $t$ as:
\begin{equation}
    \mathcal{L}_{S^2IL}^t = \frac{1}{|B|}\sum_{x \in B}\sum_{j=1}^{L_{fm}}\frac{1 - \text{SSIM}(f^t_{Lj}(x), f^{t-1}_{Lj}(x))}{2}
    \label{eq:s2il_loss}
\end{equation}
where $B$ is the mini-batch and the inner term measures the dissimilarity between the feature maps. The inner term lies in the range $[0, 1]$ since SSIM lies in the range $[-1, 1]$. Unlike \cite{Dou_2020_ECCV}, \cite{Kang_2022_CVPR}, and \cite{Bala_2024_arXiv}, we apply FD only on the last convolutional layer as it contains rich semantic and structural information. Additionally, applying distillation across all layers could hinder the model’s plasticity, as discussed in \cite{Kim_2023_CVPR} and validated by additional results presented in the supplementary material.
The overall training of task $t$ is governed by the loss:
\begin{equation}
    \mathcal{L}^t = \mathcal{L}_{cls}^t + \lambda \mathcal{L}_{S^2IL}^t
\end{equation}
where $\mathcal{L}_{cls}$ is the local similarity classifier loss adopted from \cite{Dou_2020_ECCV} and $\lambda$ is a hyperparameter adopted from \cite{Kang_2022_CVPR, Hou_2019_CVPR} that measures the degree of need to preserve old knowledge, increasing as the ratio of new to old classes grows. 
The training algorithm is listed in the supplementary material.

As an empirical proof of concept for the efficacy of $S^2IL$, we conducted the following experiment. We defined an Oracle model $O$, which is incrementally trained like $M$, except that during task $t$, $O$ has access to the entire training data of all previous tasks, rather than just exemplars, and does not use any distillation loss. We define the deviation measure of $M$ from $O$ for each class $l \in C^0$ as follows:
\begin{equation}
    D_l(M, O) = \frac{1}{L_{fm}}\sum_{j=1}^{L_{fm}}\left(1 - \frac{\alpha_{Lj, l}^{M^T} - \alpha_{Lj, l}^{M^0}}{\alpha_{Lj, l}^{O^T} - \alpha_{Lj, l}^{O^0}}\right)^2
    \label{eq:dev_oracle}
\end{equation}
where $\alpha_{Lj, l}^{M^t}$ ($\alpha_{Lj, l}^{O^t}$) is the mean Grad-CAM importance \cite{Selvaraju_2017_ICCV} of the $j^{th}$ feature map from the last convolutional layer of model $M$($O$) at task $t$, with the mean computed over the samples from class $l \in C^0$. This captures the influence of the $j^{th}$ feature map on the prediction of class $l$. Note that Grad-CAM importance is computed post-training. 
%and is not part of the incremental training task.
The fractional term in Eqn. \eqref{eq:dev_oracle} measures the evolution of the $j^{th}$ feature map of $M$ relative to $O$ across all the tasks. If $M$ behaves similarly to $O$, the fractional term will be close to $1$, resulting in a deviation score near $0$, indicating that $M$ maintains stability and plasticity like $O$. The heatmap in  Figure \ref{fig:deviation_heatmap} shows the deviations of $S^2IL$ and other models including LUCIR \cite{Hou_2019_CVPR}, PODNet \cite{Dou_2020_ECCV}, AFC \cite{Kang_2022_CVPR} and EXACFS \cite{Bala_2024_arXiv}, for a sample of base classes from CIFAR-100 under the same incremental setting as in Figure \ref{fig:importance_plot}. Clearly, $S^2IL$ exhibits the lowest deviations, demonstrating its superior stability and plasticity. Additional visual plots corresponding to other base classes are provided in the supplementary material.

%=======================================================================================================================================================================
%=======================================================================================================================================================================
%=======================================================================================================================================================================
%=======================================================================================================================================================================

\section{Results \& Discussions}
\subsection{Experimental Settings}
The SOTA benchmark datasets used to assess the performance of the proposed $S^2IL$ method are CIFAR-100 \cite{cifar100}, ImageNet-1K \cite{imagenet} and a subset version of ImageNet-1K named ImageNet-100 \cite{Rebuffi_2017_CVPR} which has $100$ randomly selected classes 

from Imagenet-1K. We provide the details of these datasets in the supplementary material. Image preprocessing and class order settings follow the methodology outlined in \cite{Dou_2020_ECCV}. We use ResNet-32 \cite{Dou_2020_ECCV} for CIFAR-100 and ResNet-18 \cite{Dou_2020_ECCV} for both ImageNet-100 and ImageNet-1K. For training, the base task (Task 0) uses half the number of total classes with the subsequent tasks taking increments of $1$, $2$, $5$, or $10$ new classes for CIFAR-100 and ImageNet-100, and $50$ or $100$ new classes for ImageNet-1K.

\begin{table*}[t!]
\centering
\begin{tabular}{|c|c|c|c|c|c|c|c|c|c|}
\hline
{\textbf{Methods}} & \multicolumn{4}{c|}{CIFAR-100} & \multicolumn{4}{c|}{ImageNet-100} \\
\cline{2-9}
& \textbf{Inc 1} & \textbf{Inc 2} & \textbf{Inc 5} & \textbf{Inc 10} & \textbf{Inc 1} & \textbf{Inc 2} & \textbf{Inc 5} & \textbf{Inc 10} \\ \hline
iCARL \cite{Rebuffi_2017_CVPR} (2017)  & 44.20±0.98 & 50.60±1.06 & 53.78±1.16 & 58.08±0.59 & 54.97 & 54.56 & 60.90 & 65.56 \\ \hline
LUCIR \cite{Hou_2019_CVPR} (2019) & 49.30±0.32 & 57.57±0.23 & 61.22±0.69 & 64.01±0.91 & 57.25 & 62.94 & 70.71 & 71.04 \\ \hline
BiC \cite{Wu_2019_CVPR}(2019) & 47.09±1.48 & 48.96±1.03 & 53.21±1.01 & 56.86±0.46 & 46.49 & 59.65 & 65.14 & 68.97 \\ \hline
PODNet \cite{Dou_2020_ECCV} (2020) & 57.86±0.38 & 60.51±0.62 & 62.78±0.78 & 64.62±0.65 & 62.48 & 68.31 & 74.33 & 75.54 \\ \hline
PODNet + AANet \cite{Liu_2021_CVPR} (2021) & - & 62.31±1.02 & 64.31±0.90 & 66.31±0.87 & - & 71.78 & 75.58 & 76.96 \\ \hline
AFC \cite{Kang_2022_CVPR} (2022) & 61.39±0.86 & 63.85±0.3 & 64.53±0.56 & 65.89±0.85 & 72.08 & 73.34 & 75.75 & 76.87 \\ \hline
EXACFS \cite{Bala_2024_arXiv} (2024) & 61.1±0.75 & 62.75±0.8 & 64.05±1.05 & 65.48±1.08 & 72.5 & 73.78 & 74.96 & 75.93 \\ \hline
eTag \cite{Huang_2024_AAAI} (2024) & - & 61.63±0.79 & 65.50 & \textbf{67.99} & - & 71.77 & 75.17 & 76.79 \\ \hline
MTD \cite{Wen_2024_CVPR} (2024) & 60.0±1.20  & 62.46±0.21 & 65.39±0.81  & 66.96±0.56 & 70.8 & 73.73 & \textbf{76.26} & \textbf{77.82} \\ \hline
$S^2IL (\textbf{Ours})$ & \textbf{62.94±1.36} & \textbf{64.23±1.24} & \textbf{65.88±0.8} & 67.35±1.15 & \textbf{73.15} & \textbf{74.27} & 75.63 & 76.52 \\ \hline
\end{tabular}
\caption{Performance Comparison using AIA (\%) on CIFAR-100 and ImageNet-100. CIFAR-100 results are averaged over 3 runs. EXACFS is implemented by us. AANet, eTag, and MTD results are from \cite{Liu_2021_CVPR}, \cite{Huang_2024_AAAI}, and \cite{Wen_2024_CVPR}, respectively. Results for other methods are based on code from \cite{Dou_2020_ECCV} and \cite{Kang_2022_CVPR}.}
\label{tab:aia_results}
\end{table*}

\begin{table}[t!]
\centering
\begin{tabular}{|c|c|c|c|}
\hline
{\textbf{Methods}} & \textbf{Inc 20} & \textbf{Inc 50} & \textbf{Inc 100} \\ \hline
AFC \cite{Kang_2022_CVPR} (2022) & - & 67.02 & 68.9 \\ \hline
MTD \cite{Wen_2024_CVPR} (2024) & 64.11 & \textbf{68.15} & \textbf{70.4} \\ \hline
$S^2IL (\textbf{Ours})$ & \textbf{64.84} & 67.23 & 68.9 \\ \hline
\end{tabular}
\caption{Performance Comparison using AIA (\%) for AFC, MTD, and S2IL on ImageNet-1K.\textsuperscript{}}
\label{tab:aia_results_imagenet1k}
\end{table}

\begin{table}[h]
\centering
\resizebox{\columnwidth}{!}{ % Auto-resize to fit within one column
\begin{tabular}{|c|c|c|c|c|c|c|c|c|}
\hline
 & iCARL & LUCIR & BiC & PODNet & AFC & EXACFS & MTD & $S^2IL$ \\
\hline
BT (\%) & -24.14 & -11.25 & -6.73 & -7.98 & -9.0 & -9.02 & -6.46 & \textbf{-5.3} \\
\hline
Fgt (\%) & 17.5 & 9.1 & 30.9 & 11.9 & 7.5 & 9.01 & 10.9 & \textbf{7.3} \\
\hline
\end{tabular}
}
\caption{Backward transfer (higher is better) and Forgetting metric (lower is better) values on the CIFAR-100 dataset with \textbf{Inc 10} setting.}
\label{tab:bwt}
\end{table}

The exemplar memory budget is fixed at $2000$ for CIFAR-100 and ImageNet-100, and at $20000$ for ImageNet-1K, with exemplar selection based on the herding technique \cite{Rebuffi_2017_CVPR}. $p$, $q$ and $r$ for SSIM are set at $0.1$, $8$ and $8$. Further details on hyperparameters are provided in the supplementary material. The performance metrics used in this study are AIA \cite{Rebuffi_2017_CVPR}, Backward Transfer (BT) \cite{Yan_2021_CVPR} and Forgetting metric (Fgt). For comparative analysis, we consider memory replay, knowledge distillation, and network expansion methods, evaluated under identical conditions.

\vspace{-3mm}
\subsection{Comparing $S^2IL$ with SOTA methods}
Table \ref{tab:aia_results} shows the results. In CIFAR-100, $S^2IL$ outperforms others, especially in the challenging \textbf{Inc 1} and \textbf{Inc 2} settings, by around 1.5\%. $S^2IL$ also performs best in the difficult \textbf{Inc 1} and \textbf{Inc 2} settings of ImageNet-100. While other distillation strategies in difficult settings like \textbf{Inc 1} prioritize stability by aligning features in magnitude and direction, they sacrifice plasticity. In contrast, $S^2IL$ maintains both stability and plasticity, excelling in such settings. The AIA of $S^2IL$ on ImageNet-1K is presented in the Table \ref{tab:aia_results_imagenet1k}
, in which we notice that $S^2IL$ performs better in the difficult INC $20$ setting while MTD is does well in INC $50$ and $100$. Results for other methods on ImageNet-1K are presented in supplementary material. The reason for $S^2IL$’s better AIA in low increment settings and MTD’s better AIA in large increment settings can be attributed to the multiple teachers in MTD that are forced to be diverse in their correct classifications, which is feasible with many classes per increment, boosting performance in large increment settings. However, with low increment settings, the room for diversity among teachers drops significantly. With no further FD in MTD, it forgets more unlike $S^2IL$. This is further highlighted by better BT and Fgt percentages for $S^2IL$ in comparison to MTD and other methods as presented in Table \ref{tab:bwt}.

\begin{table}[h]
\centering
\begin{tabular}[\columnwidth]{|c|c|c|c|c|c|c|}
\hline
\textbf{Param} & \textbf{0.1} & \textbf{0.2} & \textbf{0.4} & \textbf{4} & \textbf{8} & \textbf{16} \\ \hline
p=0, q=0, r & 60.22 & 62.01 & 64.08 & 67.17 & 67.18 & 56.7 \\ \hline
p=0, q, r=8 & 67.05 & 67.18 & 67.31 & 67.23 & 66.98 & 67.05 \\ \hline
p, q=8, r=8 & 66.92 & 66.97 & 66.68 & 53.78 & 53.9 & 53.63 \\ \hline
\end{tabular}
\caption{AIA (in \%) of $S^2IL$ for different values of $p$, $q$ and $r$ under \textbf{Inc 10} setting in CIFAR-100}
\label{tab:pqr}
\end{table}

\begin{table}[h]
\centering
\begin{tabular}{|c|c|}
\hline
\textbf{Our Model with} & \textbf{AIA (\%)} \\ \hline
only $l$  & 60.82±0.65 \\ \hline
only $c$  & 62.52±1.02 \\ \hline
only $s$  & 67.65±0.97 \\ \hline
$l$+$c$  & 64.76±0.74 \\ \hline
$l$+$s$  & 67.18±1.18 \\ \hline
$c$+$s$  & 67.59±1.17 \\ \hline
$l$+$c$+$s$  & 67.35±1.15 \\ \hline
\end{tabular}
\caption{Ablation study on components of SSIM (\textbf{Inc 10}, CIFAR-100, mean over 3 runs)} 
\label{tab:ssim_ablation}
\end{table}

\vspace{-4mm}
\subsection{Ablation Studies}
We performed an ablation study on the SSIM parameters $p$, $q$, and $r$ over the set of values $\{0.1, 0.2, 0.4, 4, 8, 16\}$ under the CIFAR-100 \textbf{Inc 10} setting. First, we optimized $r$ for the structure component, then $q$ with $r$ fixed, and finally $p$ with both $r$ and $q$ fixed. Results are shown in Table \ref{tab:pqr}. While $q=0.4$ and $q=4$ slightly outperforms $q=8$ on CIFAR-100, we selected $q=8$ as it yielded marginally better results on ImageNet-100. Additionally, we investigated the impact of the SSIM components $l$, $c$, and $s$ on the model performance. As shown in Table \ref{tab:ssim_ablation}, the absence of the structure component leads to a sharp performance drop, highlighting its importance. Despite a marginal improvement from using only structure in CIFAR-100 \textbf{Inc 10} setting, incorporating all three components ($l$, $c$, and $s$) provided better results on ImageNet-100, guiding our final choice (refer to the supplementary material). More ablation studies are presented in the supplementary material.

\vspace{-2mm}
\section{Conclusions}
FD methods in the literature force equality in the direction and magnitude among feature maps which impair the model's plasticity. The proposed method, $S^2IL$, effectively balances the stability-plasticity dilemma by enforcing structural similarity between feature maps across incremental tasks through the incorporation of SSIM. We also validated our method by testing it on SOTA benchmark datasets over the AIA, BT and Fgt metrics and find that $S^2IL$ delivers strong and comparable performance without enforcing direct feature preservation across tasks. 
\\
Computing SSIM requires either the forward propagation of exemplars through a stored copy of the old model, or retaining feature maps of all exemplars, leading to an additional memory overhead. This can pose challenges for implementation in memory-constrained settings. Recent works in CIL focus on exemplar-free approaches and it would be interesting to incorporate $S^2IL$ in such scenarios where the need for storing exemplars is eliminated.

% \section*{Acknowledgment}
% This work is dedicated to Bhagawan Sri Sathya Sai Baba, the Founder Chancellor of Sri Sathya Sai Institute of Higher Learning.

\section*{References}

\def\refname{}
\bibliography{bare_jrnl}\bibliographystyle{IEEEtran}

\clearpage
\appendix
% \documentclass{IEEEtran}

% \usepackage{amsmath}
% \usepackage{multirow}
% \usepackage{dblfloatfix}
% \usepackage{float}
% \usepackage{algorithm}
% \usepackage{algpseudocode}
% \usepackage{graphicx}
% \usepackage{caption}

% \title{$S^2IL$: Structurally Stable Incremental Learning}
% \author{Supplementary Material}

% \begin{document}

% \maketitle

% \section{Training Algorithm for the Proposed $S^2IL$ Method}

\begin{algorithm}
\caption{Training at Task $t$}
\begin{algorithmic}[1]
\State \textbf{Input:}
\State $D^t$ : Training data for task $t$
\State $E^{0 \sim t-1}$ : Set of exemplars from tasks $\{0, ..., t-1\}$
\State $p$, $q$, $r$ : SSIM params set to $0.1$, $8$, and $8$, respectively
\State $M^{t-1}$ : Trained model from task $t-1$
\State $\lambda$ : Weight for $S^2IL$ loss $\mathcal{L}_{S^2IL}$
\State \textbf{Output:} Trained model $M^t$
\State \textbf{Initialization:} $F^t \gets F^{t-1}$
\State For ($x$, $y$) $\in D^t \cup E^{0 \sim {t-1}}$
\State \quad Compute $\mathcal{L}_{S^2IL}(x)$ and $\mathcal{L}_{cls}(x)$
\State \quad $\mathcal{L}(x) = \mathcal{L}_{cls}(x) + \lambda \mathcal{L}_{S^2IL}(x)$
\State \quad Backpropagate and update model parameters
\end{algorithmic}
\end{algorithm}

\section{Datasets Details}
We utilized the following benchmark datasets: CIFAR-100, ImageNet100, and ImageNet-1K. The CIFAR-100 dataset \cite{cifar100} contains 60,000 color images across 100 classes, each with a resolution of $32 \times 32$. It is divided into 50,000 training images and 10,000 test images. ImageNet100 \cite{Rebuffi_2017_CVPR} is a subset of the original ImageNet-1K \cite{imagenet}, consisting of 100 randomly selected classes, each containing approximately 1,300 color images. The sampling method for ImageNet100 is detailed in \cite{Hou_2019_CVPR, Dou_2020_ECCV}. ImageNet-1K comprises over 1.2 million labelled training images across 1,000 object categories, with an additional 50,000 validation images. It serves as a key benchmark for evaluating machine learning models in image classification tasks.

\section{Motivation for $S^2IL$ - More Visuals}
\subsection{AFC\cite{Kim_2023_CVPR} and EXACFS\cite{Bala_2024_arXiv} Importance Values Flatten to $1$}
\begin{figure*}[h] % h = here, t = top, b = bottom, p = page
    \centering
    \includegraphics[width=0.8\textwidth]{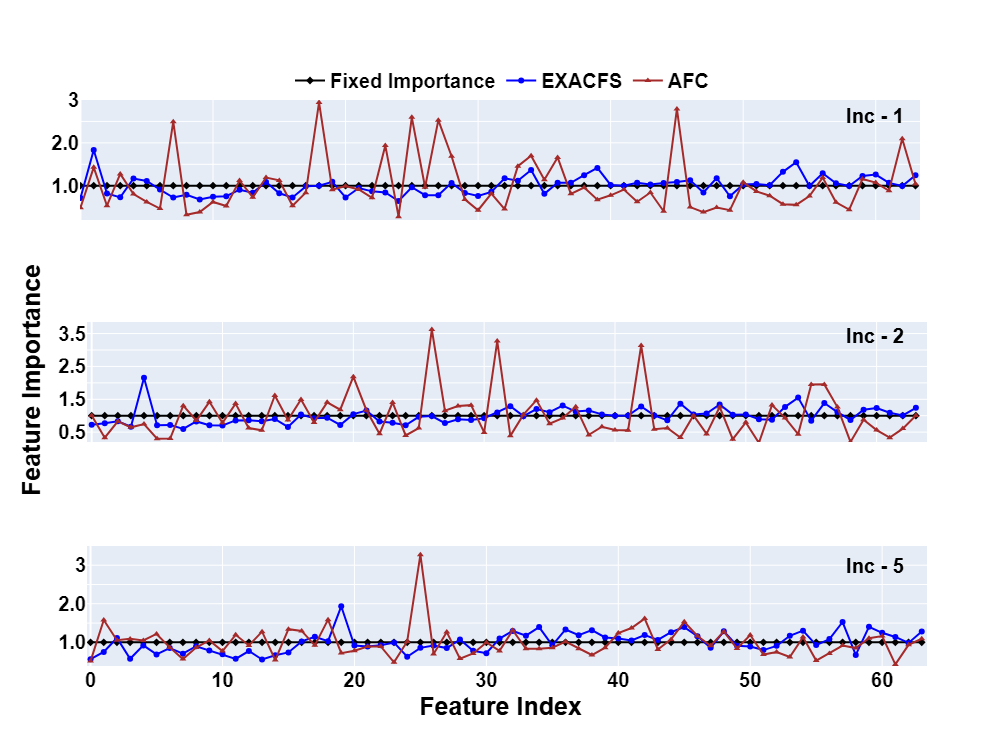} 
    \caption{Figure 1a from the manuscript extended for other incremental settings on the CIFAR-100 dataset. The importance values computed by AFC [22] and EXACFS [4] are predominantly flattened to 1.}
    \label{fig:fig1a}
\end{figure*}

We present visuals in Figure \ref{fig:fig1a} for other incremental settings than the one considered in Figure 1a in the manuscript to show that importances estimated by AFC and EXACFS flatten to $1$. Specifically, only $10\%$, $4\%$, and $1\%$ of the feature importances are significant for AFC in the \textbf{Inc 1}, \textbf{Inc 2}, and \textbf{Inc 10} settings, respectively. For EXACFS, only $1\%$ of the feature importances appear significant in all cases.

\vspace{-2mm}
\subsection{Deviations from Oracle Model}
\begin{figure*}[h] % h = here, t = top, b = bottom, p = page
    \centering
    \includegraphics[width=0.75\textwidth]{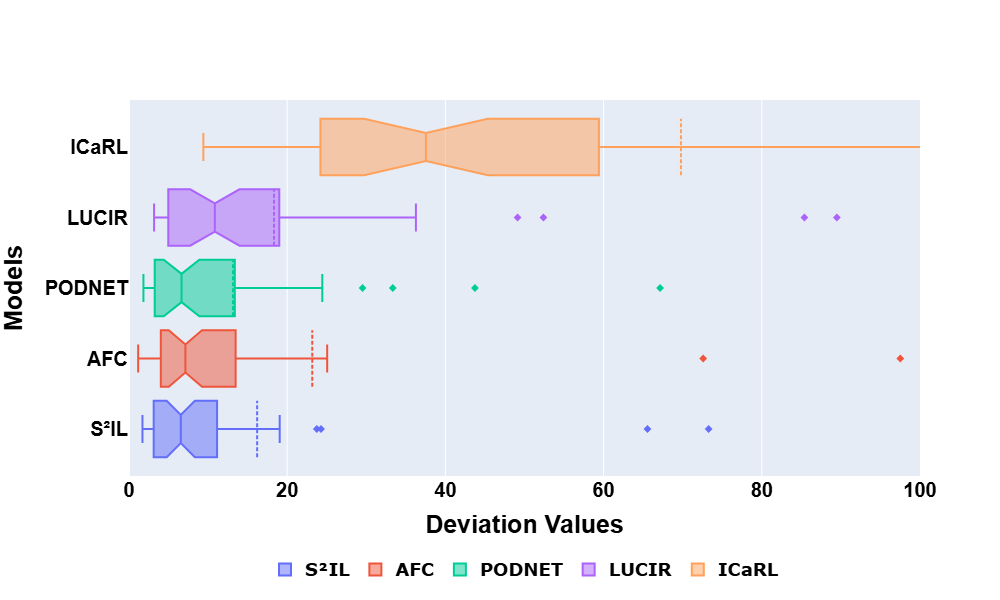} 
    \caption{Boxplot of deviations in Grad-CAM feature map importance values of various models compared to the Oracle model on the CIFAR-100 dataset}
    \label{fig:fig1b}
\end{figure*}

We present here in Figure \ref{fig:fig1b} the deviations in Grad-CAM feature map importance values of various models compared to the Oracle model on the CIFAR-100 dataset for all the $50$ base classes. Unlike the heatmap in Figure 1b (that presented deviations for a sample of classes only), which would become too large and cluttered due to all the 50 classes considered here, we use a boxplot to display these deviations, summarizing the statistics across all classes. The results clearly show that $S^2IL$ exhibits the smallest deviation from the Oracle model.

\section{Hyperparameter Details}
The model is trained on CIFAR-100, ImageNet-100, and ImageNet-1K datasets using SGD with a momentum of $0.9$. For CIFAR-100, training is conducted over $160$ epochs with a batch size of $128$, a weight decay of $0.0005$, and an initial learning rate of $0.1$, which is decayed using a \textit{CosineAnnealingScheduler}. For ImageNet-100 and ImageNet-1K, training spans $90$ epochs with a batch size of $64$, a weight decay of $0.0001$, and an initial learning rate of $0.05$, also decayed via the \textit{CosineAnnealingScheduler}. The loss function employed is the local similarity classifier loss \cite{Dou_2020_ECCV}, featuring a margin of $0.6$, a learnable scale factor initialized to $1.0$, and $10$ proxies per class. The regularization coefficient ($\lambda$) in Equation (5) is configured as $4 \times \sqrt{(|C^{0 \sim t}|/|C^t|)}$ for CIFAR-100 and $10 \times \sqrt{(|C^{0 \sim t}|/|C^t|)}$ for ImageNet datasets. Following initial training, the model undergoes fine-tuning for $20$ epochs on a balanced dataset encompassing samples from all seen classes, with learning rates of $0.05$ for CIFAR-100 and $0.02$ for both ImageNet-100 and ImageNet-1K.

\section{ImageNet-1K Results for other models}
Here we present the results for all the models considered for the comparative analysis of $S^2IL$, in Table \ref{tab:aia_results_2} for the incremental settings of Inc $50$ and Inc $100$

\begin{table}[h]
\centering
\begin{tabular}{|c|c|c|}
\hline
{\textbf{Methods}} & \multicolumn{2}{c|}
{ImageNet-1K} \\
\cline{2-3}
& \textbf{Inc 50} & \textbf{Inc 100} \\ \hline
iCARL \cite{Rebuffi_2017_CVPR} (2017) & 46.72 & 51.36 \\ \hline
LUCIR \cite{Hou_2019_CVPR} (2019) & 61.28 & 64.34 \\ \hline
BiC \cite{Wu_2019_CVPR}(2019) & 44.31 & 45.72 \\ \hline
PODNet \cite{Dou_2020_ECCV} (2020) & 64.13 & 66.95 \\ \hline
PODNet + AANet \cite{Liu_2021_CVPR} (2021) & 64.85 & 67.73 \\ \hline
AFC \cite{Kang_2022_CVPR} (2022) & 67.02 & 68.9 \\ \hline
EXACFS \cite{Bala_2024_arXiv} (2024) & - & - \\ \hline
eTag \cite{Huang_2024_AAAI} (2024) & - & - \\ \hline
MTD \cite{Wen_2024_CVPR} (2024) & \textbf{68.15} & \textbf{70.4} \\ \hline
$S^2IL (\textbf{Ours})$ & 67.23 & 68.9 \\ \hline
\end{tabular}
\caption{Performance Comparison using AIA (\%) on ImageNet-1K. EXACFS and eTag results are not available for ImageNet-1K. MTD results are from \cite{Wen_2024_CVPR}.}
\label{tab:aia_results_2}
\end{table}

\section{Additional Ablation Studies}
\subsection{Distillation across All layers vs Last layer}
The Average Incremental Accuracy of $S^2IL$, when applied on all convolution layers and when applied on last convolution layer is presented in Table \ref{tab:layers} for CIFAR-100 dataset. The significant performance gap demonstrates that applying distillation across all layers may impair the model’s plasticity.

\begin{table}[h]
    \centering
    \begin{tabular}{|c|c|c|}
        \hline
        \textbf{Incremental Setting} & \textbf{All Layers} &
        \textbf{Last Layer}\\
        \hline
        Inc 1 & 57.69 & 62.93\\
        Inc 2 & 59.72 & 64.04\\
        Inc 5 & 63.02 & 65.56\\
        Inc 10 & 65.05 & 66.74\\
        \hline
    \end{tabular}
    \caption{Average Incremental Accuracy (\%) of $S^2IL$ applied to all layers vs. last layer}
    \label{tab:layers}
\end{table}

\vspace{-3mm}
\subsection{Impact of SSIM components on $S^2IL$}
As discussed in the Ablation Studies section of the mansucript, incorporating all the three components of SSIM provides marginally better results on ImageNet-100 as shown in Table \ref{tab:only_structure}. 
The Average Incremental Accuracy of $S^2IL$ when using only the structure component for \textbf{Inc 1, 2, 5 \& 10} on ImageNet-100 is $.3\%$ to $2.5\%$ lower than when all components are included. This guides us to the choice of applying all three components of SSIM with the values $0.1$, $8.0$ and $8.0$ for the exponent components $p$, $q$ and $r$ respectively in the SSIM formula.

\begin{table}
    \centering
    \begin{tabular}{|c|c|c|}
        \hline
        \textbf{Incremental Setting} & \textbf{\textit{only s}} &
        \textbf{\textit{l + c + s}}\\
        \hline
        Inc 1 & 70.7 & 73.15\\
        Inc 2 & 71.7 & 74.27\\
        Inc 5 & 75.54 & 75.63\\
        Inc 10 & 76.18 & 76.52\\
        \hline
    \end{tabular}
    \caption{Average Incremental Accuracy (\%) of $S^2IL$ with only structure component on ImageNet-100}
    \label{tab:only_structure}
\end{table}

\subsection{Memory Ablation}
We explored two memory allocation strategies: (1) a fixed budget of 20 exemplars per class, denoted as M1, and (2) a fixed overall memory budget of 2000 exemplars, equally distributed across the previously seen classes during incremental training, denoted as M2. Table \ref{tab:mem_type} presents the results, which show a clear advantage of M2 over M1. Therefore, we adopt the M2 strategy for our model.

\begin{table}[h]
\centering
\begin{tabular}{|c|c|c|c|c|}
\hline
Type & \textbf{Inc 1} & \textbf{Inc 2} & \textbf{Inc 5} & \textbf{Inc 10} \\
\hline
M1 (CIFAR-100) & 62.93  & 64.04  & 65.56 & 66.74 \\
\hline
M1 (ImageNet-100) & 72.9 & 74.25 & 75.55 & 76.05 \\
\hline
M2 (CIFAR-100) & 62.94 & 64.23 & 65.88 & 67.35 \\
\hline
M2 (ImageNet-100) & 73.15 & 74.27 & 75.63 & 76.52 \\
\hline
\end{tabular}
\caption{Exemplar memory type study}
\label{tab:mem_type}
\end{table}

\vspace{-5mm}

% \clearpage
% \section*{References}
% \def\refname{}
% \bibliography{bare_jrnl}\bibliographystyle{IEEEtran}

% \end{document}

\end{document}